\documentclass{article}

\usepackage{microtype}
\usepackage{graphicx}
\usepackage{subfigure}
\usepackage{booktabs} % for professional tables
\usepackage{enumitem} % list params

\usepackage{wrapfig}
\usepackage{listings} % code listing
\usepackage{xcolor}
\usepackage{pgfplots}
\usepackage[justification=centering]{caption}

\definecolor{codegreen}{rgb}{0,0.6,0}

\usepackage{hyperref}

\usepackage[accepted]{mlsys2020}

\ifdefined\isaccepted
	\gdef\FB{Facebook}
\else
	\gdef\FB{Anon Inc}
\fi

\newenvironment{itemize0}[0]{\begin{itemize}[topsep=0em, itemsep=0em]}{\end{itemize}}

\mlsystitlerunning{Predictive Precompute with Recurrent Neural Networks}

\begin{document}

\twocolumn[
\mlsystitle{Predictive Precompute with Recurrent Neural Networks}

\mlsyssetsymbol{equal}{*}

\begin{mlsysauthorlist}
\mlsysauthor{Hanson Wang}{fb}
\mlsysauthor{Zehui Wang}{fb}
\mlsysauthor{Yuanyuan Ma}{fb}
\end{mlsysauthorlist}

\mlsysaffiliation{fb}{Facebook, Menlo Park, California, USA}

\mlsyscorrespondingauthor{Hanson Wang}{hansonw@fb.com}

\mlsyskeywords{
	Machine Learning, mlsys, Neural Network, Recurrent Neural Network,
	Deep Learning, Gated Recurrent Unit, Precompute, Prefetch,
	Predictive Precompute, Predictive Prefetch}

\vskip 0.3in
\begin{abstract}
In both mobile and web applications, speeding up user interface response times can often lead to significant improvements in user engagement. A common technique to improve responsiveness is to precompute data ahead of time for specific activities. However, simply precomputing data for all user and activity combinations is prohibitive at scale due to both network constraints and server-side computational costs.
It is therefore important to accurately predict per-user application usage in order to minimize wasted precomputation (\textit{``predictive precompute''}). In this paper, we describe the novel application of \textit{recurrent neural networks} (RNNs) for predictive precompute. We compare their performance with traditional machine learning models, and share findings from their large-scale production use at Facebook. We demonstrate that RNN models improve prediction accuracy, eliminate most feature engineering steps, and reduce the computational cost of serving predictions by an order of magnitude.
\end{abstract}
]

\printAffiliationsAndNotice{}  % leave blank if no need to mention equal contribution
% \printAffiliationsAndNotice{\mlsysEqualContribution} % otherwise use the standard text.

\section{Introduction}
\label{introduction}

The relationship between application latency and user engagement is well-known;
improving the responsiveness of an application
by just a few seconds can result in significant increases in user
engagement due to the limited attention span of users \cite{palmer16}.

In modern applications, the most common source of latency is data fetching.
To serve the content required for a particular activity, one might first make a network request,
retrieve the raw content from a database, and then apply some additional processing ---
each of which may take a significant amount of time.
One common strategy to improve responsiveness from the user's perspective is to \textit{precompute} the results (prefetching them ahead of time) so that it is immediately available for the user.

For any individual activity, the simplest precompute strategy is to just always perform a precompute at application startup. However, this causes issues both client-side and server-side:

\begin{itemize0}
	\item On the client side (especially on mobile clients), aggressive precompute adversely affects cellular data usage, application startup time, and battery usage, all of which negatively impact user engagement.
	\item On the server side, the computational cost of the data fetches becomes significant at scale, especially during peak hours where computational resources are relatively scarce.
\end{itemize0}

\pagebreak

To minimize computational costs, one solution is to use an approach we call
\textbf{predictive precompute}: we can predict the probability that a user
will access a particular activity given the current application state and their historical access logs.
We then only precompute data when the probability surpasses a certain threshold, significantly reducing the proportion of wasted precompute.

The key to this approach is accurately predicting user access probabilities. Estimating probabilities translates well into a standard machine learning problem, where existing research offers some solutions: for example, \cite{wang2015} describe a system to precompute links embedded in Twitter posts using linear regression on various content-specific features and \cite{Sarker2019} demonstrate the effectiveness of decision tree models for context-aware smartphone usage predictions. However, it is often difficult to ``feature engineer'' a user's historical access logs into a form that traditional machine learning methods can use.

We propose the use of deep learning models, particularly those based on \textbf{recurrent neural networks} (RNNs), as a novel improvement over previous methods. We demonstrate through offline experiments that RNNs are able to make effective use of historical data with minimal feature engineering to achieve superior prediction accuracy over traditional models. We prove these benefits in production through an online experiment where RNNs yield a \textbf{7.81\%} increase in successful precompute over a traditional model.

Finally, we highlight the benefits of the RNN computation model from a systems perspective. By eliminating the time-based aggregations used in traditional models in favor of a single hidden state, the overall computational cost of serving predictions is reduced by a factor of \textbf{10x}. 

\section{Related Work}
\label{related}

Existing literature describes relatively simple models to estimate access probabilities in the context of precompute:

\begin{itemize0}
\item \cite{wang2015} describe a system to prefetch links embedded in social network feeds using linear regression on mostly content-based features.
\item \cite{parate} describe using the CDF of the ``time since last use'' as the probability estimate.
\item \cite{Sarker2019} showcase the effectiveness of decision tree models for context-aware smartphone usage prediction.
\end{itemize0}

However, there are a few common limitations in approaches like the ones mentioned above:

\begin{itemize0}
\item Making full use of historical access logs is difficult. Most classical machine learning models operate on fixed-length feature vectors, so a common approach is to compute aggregate functions based on the timestamps of previous accesses (e.g.
time since last access, average time between historical accesses, number of accesses within a certain time window.) However, the choice of aggregations must be manually chosen through trial-and-error (``feature engineering'').
\item Incorporating contextual features from historical access logs adds another dimension of difficulty, as any time-based aggregation can be combined with any subset of context dimensions (e.g. what was the number of accesses associated with this application surface?).
\item The act of performing the predictions themselves is resource-constrained: care must be taken to ensure that model serving must not become as computationally expensive as the precomputation itself. For example, computing and serving aggregation features like the examples mentioned above may require specialized infrastructure to remain efficient at scale.
\end{itemize0}

To address these problems, we are able to draw inspiration from research in the recommendation domain, where historical user behavior is similarly predictive. In recent times, deep learning-based recommendation systems have become extremely widespread due to their effectiveness at handling complex relationships in large datasets \cite{zhang2017}.

Of particular interest are recommendation systems based on \textit{recurrent neural networks} (RNNs) due to their innate ability to model sequential data:

\begin{itemize0}

\item \cite{latentcross} describe the use of RNN-based recommender systems for video recommendations based on user actions. Of special note is the improved handling of ``contextual features'' (e.g. time, location, interface) which we find to be very relevant in the domain of predictive precompute as well.

\item \cite{soh2017} describe the use of RNNs (specifically, \textit{gated recurrent units}) to personalize recommendations in user interfaces based on a sequence of past user interactions, which is a very similar domain to the work in this paper.

\end{itemize0}

\cite{katevas2017} have also demonstrated the effectiveness of RNN architectures to predict user responses to notifications based on a sequence of mobile sensor data.

Our findings indicate that the results from the works above transfer well to the domain of predictive precompute. Using several test datasets, we compare RNN models with simpler models similar to the ones mentioned above and demonstrate the benefits of RNNs through online and offline experimentation.

\section{Defining Predictive Precompute}
\label{data}

In this paper, we will focus on precomputation of individual activities within large-scale applications, where we have datasets of access logs from previous user sessions.

To be more specific, we would like to have a function that estimates the probability that a user accesses a particular activity within an application \textit{session}, based on the current session \textit{context} and past \textit{access logs} as inputs. We will estimate the access probability at the beginning of each session and then choose to trigger precomputation at that point if the probability is greater than some fixed threshold.

\subsection{Definitions}

\textbf{Sessions} are defined as discrete time windows where the user is actively using the application; a session typically starts when the user opens the application. For simplicity, we consider each session to have a fixed length, e.g. 20 minutes, to avoid having to precisely measure when a session ends.

\textbf{Context} describes session-specific information that may be predictive of user behavior. Examples of context include:

\begin{itemize0}
\item the current timestamp (including the hour of the day, day of the week, etc.)
\item the current application surface
\item indicators visible to the user, e.g. a ``badge count'' indicating the number of unseen notifications
\end{itemize0}

\textbf{Access logs} are a sequential record of past application sessions, keyed by a user identifier. For each session we will record the \textit{context} at the start of the session, as well as an additional Boolean \textit{access flag} indicating if the activity was accessed within that application session or not. Access logs will be used as the training dataset, where access flags are used as ground truth labels and contexts are used to extract features.

\subsection{Problem Statement}
\label{problem-statement}

For any given user, assume that we have access to logged data for $n - 1$ previous user sessions, where $C_i$ denotes the context and $A_i \in \{0, 1\}$ denotes the \textit{access flag} for the $i$\textsuperscript{th} session. If we refer to the current session as session $n$, we would like to estimate the probability of an access, $P(A_{n})$, given all known information past and present:
$$
P(A_{n} \mid C_{1}, A_{1}, C_{2}, A_{2}, ..., C_{n - 1}, A_{n - 1}, C_{n})
$$

The remainder of this paper will primarily focus on how to best estimate $P(A_{n})$. We will train machine learning models treating each session as an individual data point, with the recorded value of $A_{n}$ as the ground truth (label) and $C_{1}, A_{1}, ..., C_{n - 1}, A_{n - 1}, C_{n}$ as the basis for features.

\subsubsection{Timeshifted Precompute}
\label{timeshifting}

An interesting related problem occurs in the case where we wish to enable precomputation of data \textit{prior} to the start of an application session (e.g. several hours in advance). We refer to this modified problem as \textit{timeshifted precompute}.

The primary motivation of precomputing data further in advance is to shift computational cost from peak hours to off-peak hours. At scale, being able to shift a meaningful amount of computational cost to off-peak hours smooths out the peak/off-peak power curve and can reduce overall capacity requirements.

In this scenario we do not have access to any session-specific context and instead must rely on existing access logs alone to predict the probability that an access will occur within a pre-defined \textit{peak hours} window of a particular day. If we denote access during peak hours of day $d$ as $PA_{d} \in \{0, 1\}$, then we can state the probability estimate as:

$$
P(PA_{d} \mid C_{1}, A_{1}, ..., C_{n}, A_{n})
$$

Each training example corresponds to one user $\times$ peak window pair, with the ground truth label being the presence of an access within the peak window.

\section{Datasets}
\label{datasets}

In this paper we analyze two real-world datasets where predictive precompute is being employed at \FB{}, each consisting of a random sample of $10^6$ users over 30 days. As a publically available baseline, we also make use of the \textit{Mobile Phone Use} dataset published by \cite{pielot2017}.

\subsection{Mobile Tab Access (MobileTab)}
\label{tabclick}

\begin{table}[t]
\caption{Sample data for \textit{MobileTab.}}
\label{sampledata}
\vskip 0.15in
\begin{center}
\begin{small}
\begin{sc}
\begin{tabular}{lcccr}
\toprule
Timestamp	& Access Flag	& Unread		& Active Tab \\
\midrule
1564642800  & 1			& 3			& Home \\
1564642900  & 0			& 0			& Home	 \\
1564643000  & 0        	& 1  		& Messages \\
\bottomrule
\end{tabular}
\end{sc}
\end{small}
\end{center}
\vskip -0.1in
\end{table}

Upon startup of the \FB{} mobile application, we may choose to prefetch data for certain sections (``tabs'') if we can predict that the user is likely to access them.

A \textit{session} begins when the user starts the application and ends after a fixed window of 20 minutes. For this dataset, we selected a tab with moderate usage and recorded an \textit{access} for every session where an interaction occurred with the tab within the time window.

\textit{Context} for this dataset includes the current time, unread notification count displayed over the tab icon (0-99), and the name of the active application tab at startup. \textit{Access logs} consisting of the three context features and the access flag are stored over a 30-day period.

Table \ref{sampledata} illustrates an example sequence of sessions for an individual user in the \textit{MobileTab} dataset.

\subsection{Timeshifted Data Queries (Timeshift)}
\label{searchhistory}

On the \FB{} website, data queries that are relatively static can be precomputed and cached several hours ahead of time (as described in the \textit{Timeshifted Precompute} problem statement). During off-peak hours of the day, we predict if a user will require a particular data query result in a session during peak hours on the following day.

A \textit{session} begins upon the start of a website load and ends after a fixed 20-minute window. We selected a moderately used data query and recorded an \textit{access} for every session where the data query was used.

\textit{Context} for this dataset includes only the session timestamp and a flag indicating whether or not the session occurred during peak hours. Any additional context quickly loses relevance by prediction time, where no context is available.

\subsection{Mobile Phone Use (MPU)}
\label{mpu}

\cite{pielot2017} have generously published a dataset containing data traces for 279\footnote{There are 342 users in the full dataset, but only 279 have an Android version with full support for notification tracking.} mobile phone users over the course of four weeks. We borrow heavily from the work in \cite{katevas2017} and also attempt to predict the probability that a user opens the app associated with a notification when it is received. In the context of predictive precompute, the OS could conceivably preload the app in the background for notifications with a high probability of interaction.

For \textit{MPU} we define each \textit{session} to start with the appearance of a notification, with a fixed length of 10 minutes. An \textit{access} occurs if the user opens the app associated with the notification. To provide a reasonable comparison against the previous datasets we ignore the auxiliary ``sensors'' and focus on only the access logs associated with previous notifications.

We derive four \textit{context} variables for each notification: the current time, the current screen state (on/off/unlocked), the application ID the notification was associated with, and the last opened application ID.

\subsection{Data Statistics}

Table \ref{datastatistics} displays summary statistics for each dataset. In \textit{MobileTab} and \textit{MPU} each labeled example corresponds to a single \textit{session}, while in \textit{Timeshift} an example corresponds to a single peak period (one peak period per day and 30 days per user, for a total of 30M unique training examples). Although the \textit{MPU} dataset has a very small sample of users, there is much more data per user --- on average over 8,000 notification events per user --- which still yields a sufficient amount of data for the purposes of model training.

\begin{figure}[h]
\begin{tikzpicture}
\begin{axis}[
        width=\linewidth,
        height=2.5in,
        ylabel={Percentage of users},
        xlabel={Access rate},
        xmin=0,
        xmax=1,
        ymin=0,
        ymax=1,
        legend pos=south east,
        legend style={font=\small},
]
\addlegendentry{MobileTab}
\addplot[color=red] table [x, y, col sep=comma] {cdf_tabclick.csv};
\addlegendentry{Timeshift}
\addplot[color=blue] table [x, y, col sep=comma] {cdf_timeshift.csv};
\addlegendentry{MPU}
\addplot[color=green] table [x, y, col sep=comma] {cdf_mpu.csv};
\end{axis}
\end{tikzpicture}
\caption{CDF of access rates across users.}
\label{fig:cdf}
\end{figure}
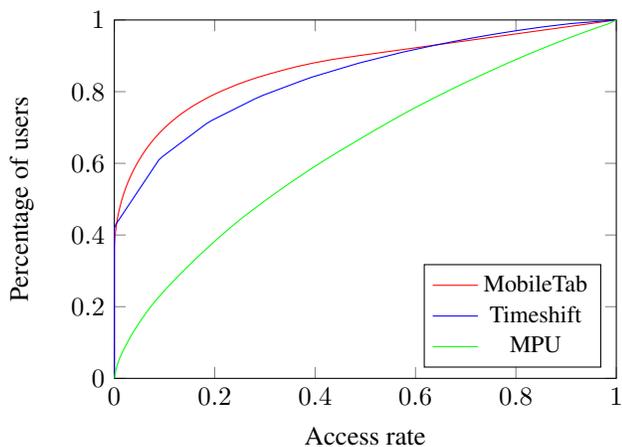

Figure \ref{fig:cdf} displays the CDF of access rates for each dataset. Note that for \textit{MobileTab} and \textit{Timeshift} a significant percentage of users (36\% and 42\% respectively) have no recorded accesses at all in 30 days; this is typical of real-world scenarios, where not all users may access a particular activity.

\begin{table}[h]
\caption{Summary of each dataset.}
\label{datastatistics}
\vskip 0.15in
\begin{center}
\begin{small}
\begin{sc}
\begin{tabular}{lcccr}
\toprule
Data set			& Positive Rate	& Sessions 	& Users \\
\midrule
MobileTab 		& 11.1\%		& 60.8M 		& 1M \\
Timeshift 		& 7.1\%		& 38.5M 		& 1M \\
MPU           	& 39.7\%		& 2.34M		& 279 \\
\bottomrule
\end{tabular}
\end{sc}
\end{small}
\end{center}
\vskip -0.1in
\end{table}

\section{Baseline Models}
\label{model}

In this section we will describe a selection of simpler machine learning models for comparison against the recurrent neural network model.

\subsection{Percentage-Based Model}
\label{percentage-based}

A very simple baseline model is to return the current access percentage based on all historical sessions for each user, disregarding additional context information. We can seed the percentage estimate for each user with the globally averaged access percentage across all sessions ($\alpha \in (0, 1)$):
$$
P(A_n) = \frac{\alpha + \sum_{i=1}^{n-1}{A_i}}{n}
$$
For \textit{Timeshift} the calculation is similar, except we average over accesses at peak rather than individual sessions:
$$
P(PA_d) = \frac{\alpha + \sum_{i=1}^{d-1}{PA_i}}{d}
$$
\subsection{Prelude: Feature Engineering}
\label{feature-engineering}

In order to make use of traditional models we must first convert all the available raw context for each session into a fixed-length numerical vector. We use common \textit{feature engineering} techniques to accomplish this.

\begin{itemize0}
\item \textbf{One-hot encoding of categorical variables.} For context variables such as the unread/notification counts, active tab, and application names, we use the standard technique of one-hot\footnote{\url{https://en.wikipedia.org/wiki/One-hot}} encoding. Note that for the tab and application name features we first limit the number of distinct values to a reasonable range by hashing and taking the remainder modulo 97.
\item \textbf{Time-based features.} Given the raw timestamp, we additionally calculate the hour of day (0 - 23) and day of week (1 - 7) and apply a one-hot encoding to them.
\item \textbf{Time-based aggregations.} We can track the number of accesses, number of sessions, and their ratio (the access percentage) for each user across a variety of time windows. In our comparisons we use the last 28 days, 7 days, 1 day, and 1 hour as time windows. We can also filter past accesses to those whose contexts match the current session context, e.g. having the same active tab or notification count (or both).  To maximize coverage, we calculate aggregations based on all (time window) $\times$ (matching subset of context) combinations.
\item \textbf{Time elapsed.} We can calculate the time difference (in seconds) from both the last access and the last session. As with the aggregation features, we also condition these to past events with a matching context subset.
\end{itemize0}

Table \ref{tbl:ablation} illustrates the importance of thorough feature engineering with GBDT models; evaluation metrics drop sharply once aggregation and time-elapsed features are removed. 

\subsection{Logistic Regression (LR)}

As a first step we use logistic regression (LR) on the features described above, treating each session as an individual data point. To give the aggregation-based features adequate warm-up time, only sessions from the latest 7 days of each dataset are used for training; of these, we take 90\% of the users as the training set and leave 10\% as a \textit{test set} for evaluation. Details are explained in Section \ref{comparison}. An additional feature pre-processing step is to bucketize \textit{time elapsed} features into 50 buckets and one-hot encode them, due to their uneven distribution. To do so, we take $\lfloor \frac{50}{15} \cdot ln(t) \rfloor$ where $t$ is the time difference in seconds; note that the largest possible $t$ (30 days) is about $e^{14.76}$ seconds.

To train a logistic regression model, we use the scikit-learn \cite{scikit-learn} \texttt{LogisticRegression}\footnote{\url{https://scikit-learn.org/stable/modules/generated/sklearn.linear_model.LogisticRegression.html}} API with the \texttt{saga} solver and default settings.

\subsection{Gradient Boosted Decision Trees (GBDT)}

Gradient boosted decision trees (GBDT) are a popular model type that provides solid results with minimal tuning. Training is similar to the logistic regression approach, but we skip the one-hot encoding step for time-elapsed features and also some categorical features (e.g. the time of day and the day of the week).

We use XGBoost\footnote{\url{https://github.com/dmlc/xgboost}} 0.90 \cite{xgboost} to train a decision tree model with mostly default settings, except for the \textit{tree depth} hyperparameter. To determine the optimal tree depth, we split off 10\% of the users from the training set as \textit{validation} set. We then use a simple exhaustive search over all possible depths in the range $[1, 10]$ to minimize the \textit{log loss} objective over the validation dataset.

With the manually engineered numerical features, we find that GBDTs are very hard to beat. We tested simple neural network architectures (e.g. a multi-layer perceptron) and could not obtain significant gains over GBDT models.

\section{Recurrent Neural Networks}

\begin{figure*}
\label{arch-diagram}
\includegraphics[width=\linewidth, keepaspectratio]{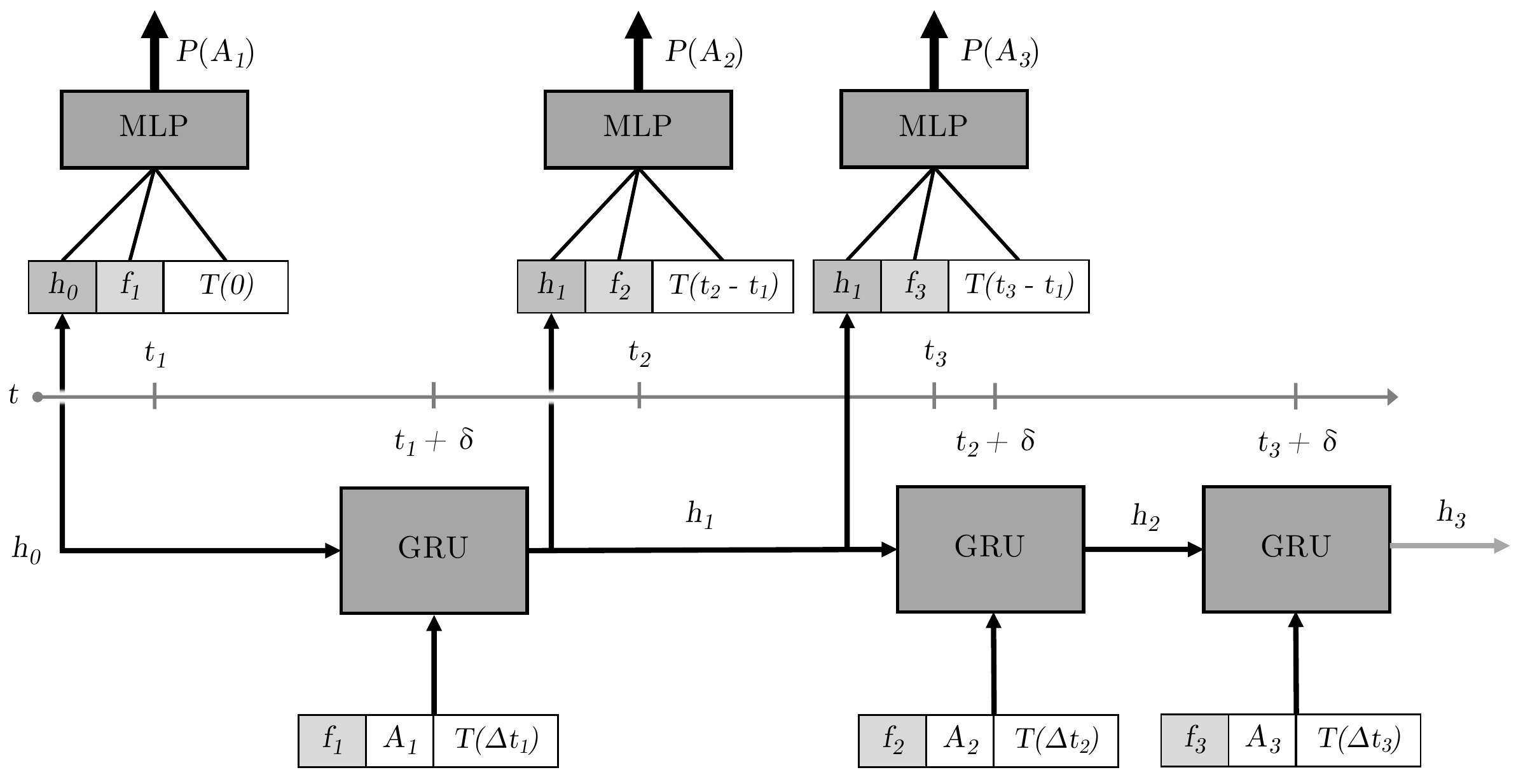}
\caption{Modeling sequences of access logs with recurrent neural networks. Multilayer perceptron ($MLP$) units produce output probabilities at time $t_i$, while hidden state updates occur through the gated recurrent units ($GRU$) at time $t_i + \delta$ due to the delay. Note that because $t_3$ occurs before $t_2 + \delta$ it cannot make use of $h_2$ and uses $h_1, t_1$ as inputs instead.}
\end{figure*}

Whereas traditional models treat the access prediction for individual sessions as independent events, the innovation of recurrent neural networks (RNNs) is to process events in a sequential manner while introducing a persistent \textit{hidden state} to carry over information from previous events.

An RNN model accepts both the current feature vector and the hidden state as inputs, and produces both a prediction and an updated hidden state for the subsequent prediction.

The primary benefit of RNN architectures is to obsolete the tedious and manual \textit{feature engineering} steps described in section \ref{feature-engineering} --- specifically the aggregation and time difference features. Instead, the RNN hidden state has the ability to automatically capture features based on past events.

Traditionally, RNN architectures have been popularized in text and audio domains, as they are well suited to handle regular sequences of information like characters or audio samples \cite{graves-speech}. In this section we describe various technical considerations when applying RNNs in the domain of \textit{predictive precompute}.

\subsection{Sequence Modeling}

Recall from section \ref{problem-statement} that for each user we have a sequence of $n$ logged historical sessions, with contexts $C_1, ..., C_n$ and access activity $A_1, ..., A_n$. Let $t_1, ..., t_n$ also denote the UNIX timestamp of each session where $t_1 < t_2 < \cdots < t_n$. To model this using RNNs, we first introduce some preliminary concepts:

\begin{itemize0}
\item \textbf{Feature extraction.} Each step of the RNN model must receive a fixed-length feature vector. While we can omit all of the aggregation features described in Section \ref{feature-engineering}, we still must construct a feature vector from each $C_i$ consisting of the one-hot encoded categorical context variables (e.g. notification count) and time-based features (hour of day, day of week). Let $f_i$ denote the feature vector of context $C_i$.
\item \textbf{Representing different time intervals.} Traditional RNN models usually operate on regular sequences where elements in the sequence represent fixed time steps or consecutive characters. However, for a sequence of user sessions, some sessions may be seconds apart, while others may be hours or days apart. To allow the network to react effectively to different timescales we input $\Delta t_i = t_{i} - t_{i-1}$ to the recurrent network at each step, where $\Delta t_1 = 0$. We find in our datasets that the distribution of $\Delta t$ tends to be power-law distributed, so we apply the log/bucketing transform described in \ref{feature-engineering} here as well, denoted $T(\Delta t_i)$.
\item \textbf{Hidden states.} Each user starts with an initial hidden state $h_0$, an all-zero vector. At the end of session $i$, the RNN model produces an updated $h_i$ based on the previous hidden state $h_{i-1}$ and inputs $f_i$, $A_i$, and $\Delta t_i$.
\item \textbf{Update delays.} To model real-world behavior accurately we must take into account two sources of delays: (1) that the ground truth $A_i$ cannot be determined until the session ends (recall that each session has a fixed length, e.g. 20 minutes), and (2) that obtaining $h_i$ is not instantaneous (i.e. it takes some time, $\epsilon$). To address this we define a \textit{lag parameter}, $\delta$, equal to the session length plus $\epsilon$.
\item \textbf{Functions for hidden updates and predictions.} Abstractly, an RNN can be separated into two functions: an updater $RNN_{update}$ which produces new hidden states, and a feed-forward network $RNN_{predict}$ which produces predictions as output. It is often convenient to combine the two into a single model that produces both outputs simultaneously, but this separation is required in order to properly model the lag $\delta$.
\end{itemize0}

Putting everything together, we can define a sequence of hidden states $h_0 = \mathbf{0}, h_1, ..., h_n$, with a recurrence relation:
\begin{equation}
h_i = RNN_{update}(h_{i-1}, [f_i; A_i; T(\Delta t_i)])
\end{equation}
To obtain a prediction for $P(A_i)$, we use $RNN_{predict}$ with the \textit{latest known} hidden vector accounting for update lag, denoted $h_k$, where $k$ is the maximum $k$ such that $t_k < t_i - \delta$ (if no such $k$ exists, then we let $k = 0$ and $t_i - t_k = 0$):
\begin{equation}
P(A_i) = RNN_{predict}(h_k, [f_i; T(t_i - t_k)])
\end{equation}

For \textit{timeshifted precompute} (\ref{timeshifting}), predictions do not have access to $f$ or $t$ and instead can only use $start_d$ and $h_k$, where $start_d$ marks the start of the peak period on day $d$ and $k$ is the maximum index such that $t_k < p_1 - \delta$:
\begin{equation}
P(PA_d) = RNN_{predict}(h_k, [T(start_d - t_k)])
\end{equation}
Intuitively, we can think of the hidden vectors $h$ as an encoding of the sequences of contexts $C$, accesses $A$, and timestamps $t$. Many of the feature engineering techniques in section \ref{feature-engineering} can be seen as attempts to do this in a manual way. However, through training a recurrent neural net we hope to \textit{learn} the optimal way to encode the entire sequence into a single vector rather than relying on manual tuning and heuristics. Another significant benefit is the \textit{incremental} nature of hidden updates: with manual feature engineering we need to store and retrieve the entire sequence, but with RNNs we only need the last known hidden vector.

\subsection{Model Architecture}

A few options are available for hidden state updates ($RNN_{update}$). For this paper, we evaluated three options: a basic $tanh$-based recurrent unit, a gated recurrent unit (GRU) and a long short-term memory (LSTM) unit. \cite{chung14} found that GRU and LSTM units result in comparable performance on a number of example datasets, while $tanh$ performance lags behind.

Empirically we find that GRUs provide the best performance over all of the datasets (at least without significant tuning).

The primary hyperparameter available when using RNN units is the dimensionality of the hidden vectors. Empirically, $d = 128$ seems to be a good dimensionality for all datasets. Another possible modification is to stack multiple GRU or LSTM units on top of each other; however, we report similar findings to \cite{latentcross}, where the addition of multiple GRU units did not provide a meaningful improvement over a single unit.

For $RNN_{predict}$, a simple architecture where the input vector and hidden vector are concatenated and passed into a feed-forward multilayer perceptron (MLP) provides good performance. Inspired by \cite{latentcross}, we find that an element-wise multiplication of the hidden vector with a \textit{latent factor} derived from the context provides a meaningful improvement:
$$
h'_{i} = h_{k} \circ (1 + L([f_i; T(t_i - t_k)]))
$$
where $k$ is the latest known index as described previously and $L$ is a linear transformation matrix.

As for the MLP layer, we find that a single hidden layer with 128 neurons combined with a rectified linear unit (ReLU) layer seems to be sufficient for the best performance; adding more layers does not lead to meaningful improvements. We can therefore summarize the formulation as:
$$
P(A_i) = \sigma(b_2 + W_2 \cdot ReLU(b_1 + W_1 [h'_i; f_i; T(t_i - t_k)]))
$$
Here $\sigma$ denotes the sigmoid function while $b_1, b_2$ represent constant bias vectors and $W_1, W_2$ represent linear transformation matrices.

\subsection{Loss Functions}

For binary classification problems, \textit{log loss} is the standard loss measurement function. For an individual access prediction $P(A_i)$ the log loss is defined as:
$$
-[A_i \cdot log(P(A_i)) + (1 - A_i) \cdot log(1 - P(A_i))]
$$
For traditional tasks, training is often optimized over the \textit{log loss} averaged over all points in the training dataset. However, we find that this is suboptimal when comparing evaluation metrics over later days; it over-weights errors from predictions early in the sequence (when only a small number of access logs have been included in the hidden state). On the other hand, optimizing directly on later days (e.g. the last 7 days) appears to be suboptimal as well, possibly because the gradient is less stable for later elements in the sequence\footnote{\url{https://en.wikipedia.org/wiki/Vanishing_gradient_problem}}. Empirically, we consistently find it is best to train on the log loss for the last 21 days out of the 30 days available for each user. We did not find weighing the loss with an exponential time decay to be of significant benefit.

Each session is weighted equally, which does mean that users with more active sessions have greater influence over the model's performance. Although this is desirable when considering computational costs as a whole, different weighting schemes can be explored if we wish to give more weight to inactive users as well.

\section{RNN Training}

RNN models are trained using \textit{PyTorch} v1.1\footnote{\url{https://github.com/pytorch/pytorch/releases/tag/v1.1.0}} using the \textit{Adam} optimizer with a learning rate of $1\mathrm{e}{-3}$. We also include a dropout layer in the middle of the MLP set to 20\% to prevent overfitting. Figure \ref{pytorch} displays sample PyTorch code for the key model definitions.

Each dataset is randomly split into training and test groups \textit{by user}, with 90\% of users in the training group and 10\% of users in the test group. We opted for a user-based split rather than a time-based split due to the limited number of days available; empirically this did not seem to introduce data leakage (validated through online results). Due to the small number of users in the \textit{MPU} dataset, we used a  \textit{k-fold cross-validation} setup with $k = 4$ and trained a separate model on each split. Evaluation metrics are measured over the combined cross-validated predictions (from all 4 folds).

\begin{figure}
\begin{lstlisting}[language=Python]
import torch.nn as nn

class RNNClassifier(nn.Module):

	# i_n = feature vector dimensions
	# h_n = hidden dimensions
	# w_n = number of hidden neurons
	def __init__(self, i_n, h_n, w_n):
	  self.L = nn.Linear(i_n, h_n)
	  self.W_1 = nn.Linear(i_n + h_n, w_n)
	  self.W_2 = nn.Linear(w_n, 1)
	  self.Dropout = nn.Dropout(0.2)  
	  self.GRU = nn.GRUCell(i_n + 1, h_n)
	
	# Returns a new hidden vector h_{i+1}.
	# f_i includes both the feature vector
	# and the encoded time difference.
	def hidden_forward(self, h_i, f_i, A_i):
	  return self.GRU(
	  	torch.cat((f_i, A_i), 1), 
	  	h_i,
	  )
	  
	# Predicts P(A_{i+1}).
	def forward(self, h_k, f_i):
	  cross_h_i = h_k * (1 + self.L(f_i))
	  mlp = self.W_1(
	    torch.cat((cross_h_i, f_i), 1),
	  )
	  mlp = torch.relu(self.Dropout(mlp))
	  return torch.sigmoid(self.W_2(mlp))
\end{lstlisting}
\caption{Sample PyTorch code for key model definitions.}
\label{pytorch}
\end{figure}
\subsection{Minibatch Training}
\label{minibatch}

For \textit{MobileTab} and \textit{Timeshift} we can significantly increase training speed by using \textit{minibatch} training with batches of 10 users. Note that for \textit{MPU} this becomes ineffective due to the small number of users versus high number of sessions per user, and we fall back to processing users individually.

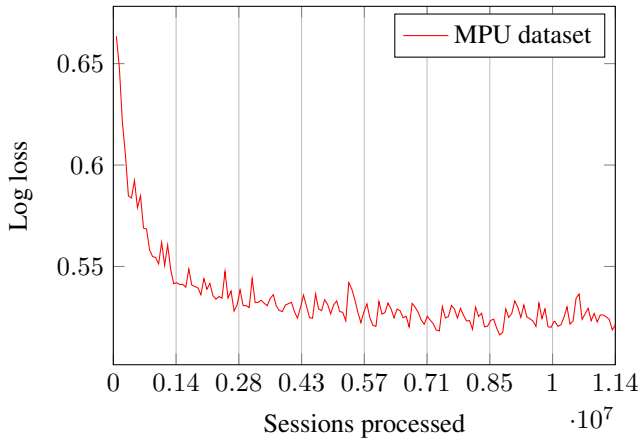
\begin{figure}[t]
\begin{tikzpicture}
\begin{axis}[
        width=\linewidth,
        height=2.5in,
        ylabel={Log loss},
        xlabel={Sessions processed},
        xmin=0,
        xmax=11390763,
        xtick distance={1423845.375},
        xmajorgrids=true,
]
\addlegendentry{MPU dataset}
\addplot[color=red] table [x, y, col sep=comma] {mpu_log_loss.csv};
\end{axis}
\end{tikzpicture}
\caption{Log loss vs. number of sessions processed (based on the full cross-validated dataset). Each vertical line represents the end of one epoch, with 8 epochs in total.}
\label{learningcurves}
\end{figure}

For each minibatch we compute predictions for the last 21 days and then calculate the average log loss over all prediction/label pairs. The loss gradient of each minibatch is then back-propagated to complete one training iteration. For the larger datasets, training converges in just one epoch, but for the \textit{MPU} dataset a total of 8 epochs are required for convergence. Figure \ref{learningcurves} displays the training log loss curves.

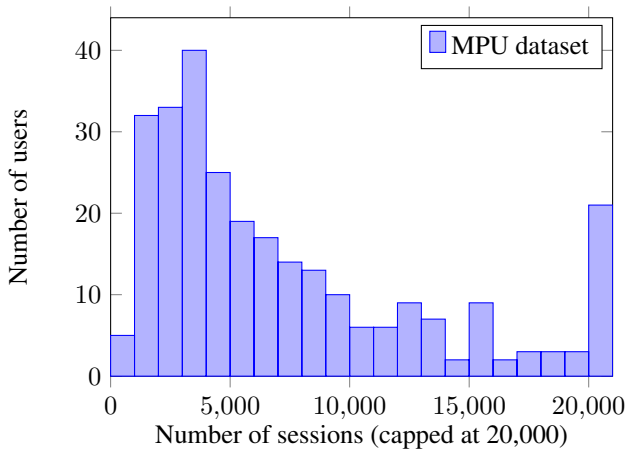
\begin{figure}
\begin{tikzpicture}
\begin{axis}[
        ybar,
        width=\linewidth,
        height=2.5in,
        ylabel={Number of users},
        xlabel={Number of sessions  (capped at 20,000)},
        scaled x ticks = false,  
        ymin=0,
        xmin=0,
        xmax=21000,
]
\addlegendentry{MPU dataset}
\addplot+[ybar interval,mark=no] table [x, y, col sep=comma] {mpu_dist.csv};
\end{axis}
\end{tikzpicture}

\caption{Distribution of \textit{MPU} session counts.}
\label{fig:mpu-dist}

\end{figure}

We find that a key optimization to speed up training is to evaluate minibatches via custom parallelism. While a standard approach is to pad user histories to a uniform length before evaluating them in batch, in practice the distribution of access history lengths has a very long tail (Figure \ref{fig:mpu-dist}). This results in an excessive amount of operations wasted on padding values. Instead, we can evaluate predictions and calculate gradients for each user on a separate thread and then accumulate the gradients afterwards. Models train twice as quickly with this approach versus the padded batch approach.

A final consideration for the \textit{MPU} dataset is to truncate user histories to the most recent 10,000 sessions. This limits the effect of long tail users on training time without having a noticeable impact on model quality.

\section{Evaluation Results}
\label{comparison}

Unless otherwise noted, evaluation metrics are reported on the \textit{test} dataset for each model. We are careful to use the same train / test split for all models (for \textit{MPU}, the same 4-fold validation sets). We split based on users because it allows more historical data per user at training time. Another key consideration is the timeframe used for evaluation: evaluating on the full 30-day period does not accurately reflect real-world performance because the majority of users already have a full 30 days of history. For example, on any given day in the \textit{MobileTab} dataset, less than 1\% of sessions have no previous history in the previous 29 days. In practice, only 1\% or fewer users do not already have previous logged history over the past 30 days. Therefore, we evaluate predictions on \textit{the last 7 days} of testing data in each dataset to get a better estimate of performance in production.

\begin{figure}[h]
\vskip -0.1in
\begin{tikzpicture}
\begin{axis}[
        width=\linewidth,
        height=2.5in,
        ylabel={Precision},
        xlabel={Recall},
        xmin=0,
        xmax=1,
        ymax=1,
        xmajorgrids=true,
	    ymajorgrids=true,
        legend style={font=\small},
]
\addlegendentry{\%Based}
\addplot[color=orange] table [x, y, col sep=comma] {tabclick_pct_pr.csv};
\addlegendentry{LR}
\addplot[color=red] table [x, y, col sep=comma] {tabclick_lr_pr.csv};
\addlegendentry{GBDT}
\addplot[color=green] table [x, y, col sep=comma] {tabclick_gbdt_pr.csv};
\addlegendentry{RNN}
\addplot[color=blue] table [x, y, col sep=comma] {tabclick_rnn_pr.csv};
\end{axis}
\end{tikzpicture}
\vskip -0.1in
\caption{Precision-recall curve comparison for \textit{MobileTab}.}
\label{fig:prcurves}
\end{figure}
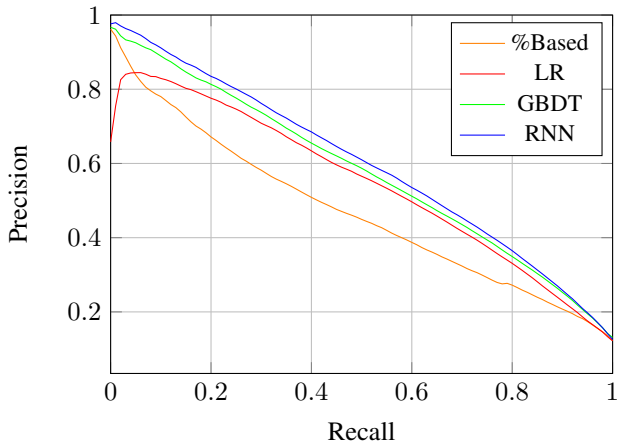

As for the evaluation metric itself, the most important metrics in the predictive precompute domain are \textit{precision} and \textit{recall}. \textit{Precision} corresponds to the percentage of precomputations that were followed by an actual access, while \textit{recall} corresponds to the percentage of accesses that were successfully precomputed. In practice, improvements in recall are almost linearly correlated to reductions in application latency. Figure \ref{fig:prcurves} shows the full precision-recall curve\footnote{As calculated via \url{https://scikit-learn.org/stable/modules/generated/sklearn.metrics.precision_recall_curve.html}} across all tested models for \textit{MobileTab}.

\begin{table}[h]

\caption{Comparison of PR-AUC values. The improvement percentage is calculated relative to the GBDT PR-AUC.}
\label{tbl:pr-auc}
\begin{center}
\begin{small}
\begin{sc}
\begin{tabular}{lcccr}
\toprule
Model				& MobileTab	& Timeshift		& MPU \\
\midrule
PercentageBased		& 0.470		& 0.260		& 0.591 \\
LR  					& 0.546		& 0.290		& 0.683 \\
GBDT  				& 0.578      	& 0.311 		& 0.686 \\
RNN					& \textbf{0.596}		& \textbf{0.335}		& \textbf{0.767} \\
\midrule
Improvement		& 3.11\%		& 7.72\%		& 11.8\%  \\ 
\bottomrule
\end{tabular}
\end{sc}
\end{small}
\end{center}

\caption{Comparison of recalls at 50\% precision.}
\label{tbl:recall0.5}
\begin{center}
\begin{small}
\begin{sc}
\begin{tabular}{lcccr}
\toprule
Model				& MobileTab	& Timeshift		& MPU \\
\midrule
PercentageBased		& 0.413		& 0.124		& 0.811 \\
LR  					& 0.596		& 0.153		& 0.906 \\
GBDT  				& 0.616      	& 0.176		& 0.917 \\
RNN					& \textbf{0.642}		& \textbf{0.209}		& \textbf{0.977} \\
\midrule
Improvement       & 4.22\%		& 18.8\%		& 6.54\% \\
\bottomrule
\end{tabular}
\end{sc}
\end{small}
\end{center}

\caption{Ablation study of feature engineering on GBDT models on the \textit{MPU} dataset. A: time-based aggregations, E: time elapsed features, C: contextual features}
\label{tbl:ablation}
\begin{center}
\begin{small}
\begin{sc}
\begin{tabular}{lccr}
\toprule
Features & PR-AUC & Recall@50\%   \\
\midrule
C  & 0.588  & 0.848 \\
E + C & 0.642 & 0.883 \\
A + E + C  & 0.686     & 0.917    \\
\midrule
RNN & 0.767 & 0.977 \\
\bottomrule
\end{tabular}
\end{sc}
\end{small}
\end{center}
\vskip -0.1in
\end{table}

To obtain a single numerical metric for model comparison we use the \textit{area under the precision-recall curve} (PR-AUC): \cite{prroc} show that PR-AUC tends to be an effective measure when dealing with highly skewed datasets (\textit{MobileTab} and \textit{Timeshift}). Table \ref{tbl:pr-auc} compares the PR-AUC across all tested models and datasets. When applying models in practice, we typically select a threshold that keeps wasted precomputations within an acceptable range (i.e. maximizing \textit{recall} while constraining on \textit{precision}; for example constraining precision to 50\%). Table \ref{tbl:recall0.5} compares the recall for each model at a fixed 50\% precision, where the difference between models becomes more apparent for \textit{MobileTab} and \textit{Timeshift}.

\section{Online Experimentation}
\label{production}

While the offline experiments described above show clear improvements, we also have results from online experiments to verify that they carry over to production environments. For the \textit{MobileTab} dataset, we productionized the RNN model to replace an existing production GBDT model as follows:

\begin{itemize0}
\item The most recent hidden state for each user (a 128-element floating point vector) and session timestamp are stored in a \textit{real-time data store} similar to \textit{Redis}\footnote{\url{https://redis.io}}.
\item TorchScript\footnote{\url{https://pytorch.org/docs/stable/jit.html}} versions of the \textit{MLP} and \textit{GRU} models are made available in a remote execution environment.
\item At session startup time, the most recent hidden state along with the current context variables are retrieved and sent through the \textit{MLP} part of the model to calculate an access probability $p$. We eagerly precompute and retrieve the tab contents if $p$ is greater than a fixed threshold, chosen to target a precision of 60\%. This corresponds to a recall of about 51.1\% in the RNN model vs. 47.4\% in the GBDT model. This comes out to a \textbf{7.81\%} increase in ``successful prefetches'' (i.e. accesses that were successfully prefetched).
\item Context variables are sent to a \textit{stream processing} system similar to \textit{Apache Kafka}\footnote{\url{https://kafka.apache.org}}, tagged by a unique session ID. Tab accesses are also sent to the same system with a matching session ID. Events are buffered by session ID, and after a timer corresponding to the session length fires, the context $C_i$ and access flag $A_i$ are computed. We then retrieve the most recent hidden state  for the user $h_i$ and execute the \textit{GRU} part of the model to calculate and store a new hidden state.
\end{itemize0}

We report several observations after monitoring the behavior of the productionized model over a period of about 90 days:

\textbf{Relative production resources.} RNN models are indeed more resource intensive --- empirically the TorchScript model is about 9.5x more computationally intensive than a GBDT model. However, in practice, the most compute-intensive component is actually the serving of aggregate access percentages and time elapsed features, which requires about two orders of magnitude more compute than the model computation itself. One approach is to retrieve the 30-day access logs for each user to compute aggregations on the fly, but some users have hundreds or thousands of past accesses which makes this impractical. Instead, aggregations are computed using a stream processing service in combination with a key-value store. However, we still need to keep track of every combination of context values in order to serve context-dependent aggregations, which may result in thousands of unique keys per user. For example, \textit{MobileTab} requires about 20 aggregation feature lookups for every individual prediction.

In contrast, with the RNN model we only need to make one key-value lookup to retrieve a 128-dimensional (512-byte) hidden vector for each prediction. By decreasing both the storage footprint and request volume, this reduces the overall serving computational cost by about \textbf{10x} in practice. Furthermore, if necessary, hidden states allow for very fine-grained control over resource usage via the hidden state dimensionality. In more resource constrained environments, we can easily train a model that has fewer hidden dimensions to trade off model quality for a smaller storage footprint per user. Neural network quantization methods can also be applied to store single bytes instead of floating-point numbers for each dimension.

\begin{figure}[t]
\begin{tikzpicture}
\begin{axis}[
        width=\linewidth,
        height=2.5in,
        ylabel={PR-AUC},
        xlabel={Days since experiment start},
        xmin=0,
        xmax=30,
        legend pos=south east,
        legend style={font=\small},
]
\addlegendentry{RNN}
\addplot[color=blue] table [x, y, col sep=comma] {tabclick_prod_pr.csv};
\addlegendentry{GBDT}
\addplot[dashed, color=blue] table [x, y, col sep=comma] {tabclick_prod_gbdt_pr.csv};
\end{axis}
\end{tikzpicture}
\caption{Online PR-AUC for \textit{MobileTab}.}
\label{fig:prod-pr}
\end{figure}
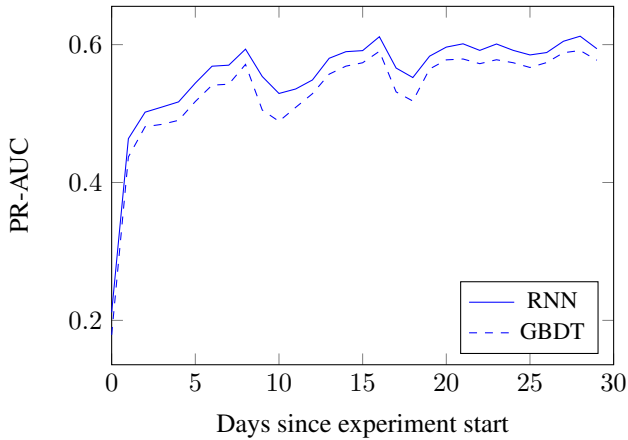

\textbf{Cold start behavior.} In our online experiment, we compared two groups of users starting with an empty history to compare the warmup behavior between the GBDT and RNN models. We find that it takes about 14 days for the RNN model to stabilize, and that it is consistently superior than the GBDT model. Figure \ref{fig:prod-pr} displays the online PR-AUC for the first 30 days of the online experiment.

\textbf{Long-term model quality and stability.} Despite the fact that the training data only spans 30 days, we see that the empirical precision and recall are consistent with the results obtained from offline experiments, and continue to maintain the same level of quality (with no sign of degradation) over a 90 day period. This suggests that the hidden states produced by the RNNs are stable over long-term periods.

\textbf{Retraining the model.} While not tested in production, it should be possible to train newer versions of the RNN model using the existing hidden states as the value for $h_0$, thus providing a path to replacing the production model without invalidating all existing hidden states. Another approach may be to preserve the \textit{GRU} parameters and hidden states and retrain only the \textit{MLP} portion of the model, which is significantly faster to retrain.

\textbf{Tradeoffs.} While RNNs have a clear advantage in model performance and significantly reduce serving computational costs, the primary drawbacks are 1) the increased training time and 2) the increased amount of data required to train an effective model. The \textit{MPU} dataset is a realistic baseline for the amount of data required, with $2 \times 10^6$ sessions, and takes about 10 hours to complete 8 training epochs. In contrast, the aggregation-based features described in Section \ref{feature-engineering} are very generically applicable to any predictive precompute use case, and GBDT models can be trained in minutes with just $10^4$ data points. Finally, hidden states are almost entirely ``black-box'' and are not easily explainable, whereas aggregation functions are easily human-interpretable.

\section{Conclusion}
\label{conclusion}

We present a review of existing techniques that can be applied to \textit{predictive precompute} problems as well as a selection of real-world datasets for comparison.

We demonstrate the novel use of recurrent neural network (RNN) models to achieve state-of-the-art results in this domain. In addition to achieving superior precision and recall metrics, RNNs significantly reduce the need for manual feature engineering due to the automatic encoding of historical information into hidden states.

We show that these advantages carry over to an online production environment, where models maintain consistent performance over an extended 90-day period.
We highlight how RNN models can help decrease the computational cost of serving models by an order of magnitude by encoding all prior history into a compact hidden state.

In closing, we hope that the techniques described in this paper make it easier for other applications to utilize predictive precompute.

\subsection{Future Work}

\textbf{Reusable models.} The very simple \textit{percentage-based} model described in \ref{percentage-based} in some sense acts as a ``universal model'' that works as a solid baseline across all use cases with almost no training. In a similar vein it may be possible to create a generic RNN-based model that uses only past session timestamps and their access labels to produce high-quality estimates without any pre-training.

\textbf{Interpretable hidden states.} The hidden state model is practically a black box. Extracting interpretable relations from the hidden state could suggest ways of feature engineering the dataset to enable the use of simpler models.

\bibliographystyle{mlsys2019}
\bibliography{paper}

%%%%%%%%%%%%%%%%%%%%%%%%%%%%%%%%%%%%%%%%%%%%%%%%%%%%%%%%%%%%%%%%%%%%%%%%%%%%%%%
%%%%%%%%%%%%%%%%%%%%%%%%%%%%%%%%%%%%%%%%%%%%%%%%%%%%%%%%%%%%%%%%%%%%%%%%%%%%%%%
% SUPPLEMENTAL CONTENT AS APPENDIX AFTER REFERENCES
%%%%%%%%%%%%%%%%%%%%%%%%%%%%%%%%%%%%%%%%%%%%%%%%%%%%%%%%%%%%%%%%%%%%%%%%%%%%%%%
%%%%%%%%%%%%%%%%%%%%%%%%%%%%%%%%%%%%%%%%%%%%%%%%%%%%%%%%%%%%%%%%%%%%%%%%%%%%%%%
% \appendix
% \section{Please add supplemental material as appendix here}
%
% Put anything that you might normally include after the references as an appendix here, {\it not in a separate supplementary file}. Upload your final camera-ready as a single pdf, including all appendices.

%%%%%%%%%%%%%%%%%%%%%%%%%%%%%%%%%%%%%%%%%%%%%%%%%%%%%%%%%%%%%%%%%%%%%%%%%%%%%%%
%%%%%%%%%%%%%%%%%%%%%%%%%%%%%%%%%%%%%%%%%%%%%%%%%%%%%%%%%%%%%%%%%%%%%%%%%%%%%%%

\end{document}